%% file: main.tex
\pdfoutput=1

\documentclass[11pt]{article}

\usepackage[final]{acl}

\usepackage{times}
\usepackage{latexsym}

\usepackage[T1]{fontenc}

\usepackage[utf8]{inputenc}

\usepackage{microtype}

\usepackage{inconsolata}

\usepackage{graphicx}
\usepackage{subcaption}

\usepackage{amsmath}
\usepackage{amsfonts}
\usepackage{mathtools}
\usepackage{algpseudocode}
\usepackage{algorithm}
\usepackage{booktabs} 
\usepackage{xcolor}
\usepackage{enumitem}

\title{Flexible-length Text Infilling for Discrete Diffusion Models}

\author{Andrew Zhang \qquad Anushka Sivakumar \qquad Chiawei Tang \qquad Chris Thomas \\
Department of Computer Science\\
    Virginia Tech \\
    {\tt \{azhang42,anushkas01,cwtang,christhomas\}@vt.edu}
}

\begin{document}
\maketitle
\begin{abstract}
Discrete diffusion models are a new class of text generation models that offer advantages such as bidirectional context, parallelizable generation, and flexible prompting compared to autoregressive models. However, a critical limitation has been the inability to perform flexible-length or flexible-position text infilling without access to ground-truth positional data. We introduce \textbf{DDOT}\footnote{Project page: \url{https://andrew-zhang.github.io/ddot-page}} (\textbf{D}iscrete \textbf{D}iffusion with \textbf{O}ptimal \textbf{T}ransport Position Coupling), a discrete diffusion model that overcomes this limitation by \emph{jointly} denoising token values and token positions using a novel sample-level optimal transport coupling. This coupling preserves relative token ordering while dynamically adjusting the positions and lengths of infilled segments. DDOT is orthogonal to existing discrete text diffusion methods and is compatible with various pretrained text denoisers. On text-infilling benchmarks such as One-Billion-Word and Yelp, DDOT outperforms naive diffusion baselines and achieves performance on par with state-of-the-art non-autoregressive models, while improving training efficiency and prompting flexibility.
\end{abstract}

\section{Introduction}
While autoregressive (AR) models are effective, they also involve highly sequential sampling, cannot use bidirectional context, and constrain architectures by requiring a decoder mask. In contrast, discrete diffusion models \cite{lou2024discrete, sahoo2024simpleeffectivemaskeddiffusion} can parallelize generation by denoising multiple tokens simultaneously, use bidirectional context, do not need a decoder mask, and allow for more controllable generation. Previous work \cite{gat2024discrete, shi2024simplified} has demonstrated that discrete diffusion models can handle prompts at arbitrary locations, whereas autoregressive models are only capable of left-to-right text completion. However, this advantage has a significant limitation: existing diffusion models cannot alter the distances between these prompt tokens. Consequently, existing text diffusion models cannot generate the ground-truth sample without access to the oracle positions of the prompt and infilled text.

\begin{figure}[t]
    \centering
    \includegraphics[width=1\linewidth]{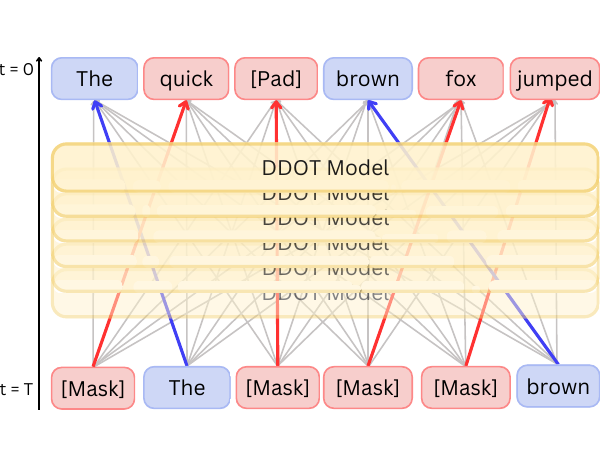}
    \vspace{-2em}
    \caption{Our diffusion across token positions enables dynamic token movement for infilling. Unlike prior methods, DDOT learns to move masked tokens to appropriate locations, such as to the right of ``brown,'' even if that position was not initially masked. The OT coupling (colored lines) simplifies this learning by drastically reducing the set of possible permutations.}
    \vspace{-1em}
    \label{fig:concept}
\end{figure}

We solve this issue by enabling discrete diffusion models to \textit{learn where to move tokens}. Specifically, we design a diffusion process that operates across token positions (in addition to token values), allowing the model to vary the positions and lengths of infilled spans (\autoref{fig:concept}). Furthermore, given the importance of token positioning in preserving semantic meaning \cite{he2020deberta}, we incorporate sample-level OT coupling \cite{tong2024improving} to maintain relative token ordering throughout the diffusion process. Even minor positional changes can dramatically alter meaning, as seen in phrases like ``The child's green coat'' and ``The green child's coat.'' DDOT’s OT coupling preserves this relative ordering throughout the diffusion process while supporting flexible-length text infilling. Our OT coupling prevents such swaps and drastically improves DDOT's training efficiency and downstream performance across all studied benchmarks and metrics. Extensive experiments show that DDOT outperforms naive diffusion baselines and achieves performance on par with state-of-the-art non-autoregressive (NAR) models.

In summary, our contributions are as follows:
\begin{itemize}
    \item We propose \textbf{DDOT}, the first discrete text diffusion method for infilling arbitrary text sequences without ground-truth span lengths.
    \item We provide \textbf{extensive experiments} on DDOT that show it outperforms diffusion baselines and achieves performance on par with state-of-the-art NAR models.
    \item We provide detailed ablations and visualizations that \textbf{verify DDOT's effectiveness} in adjusting the positions and lengths of infilled text spans and provide insights into our novel sample-level OT coupling. The OT coupling significantly outperforms naive diffusion baselines across all tested benchmarks and metrics.
\end{itemize}

\begin{figure*}[t]
    \centering \includegraphics[width=\linewidth]{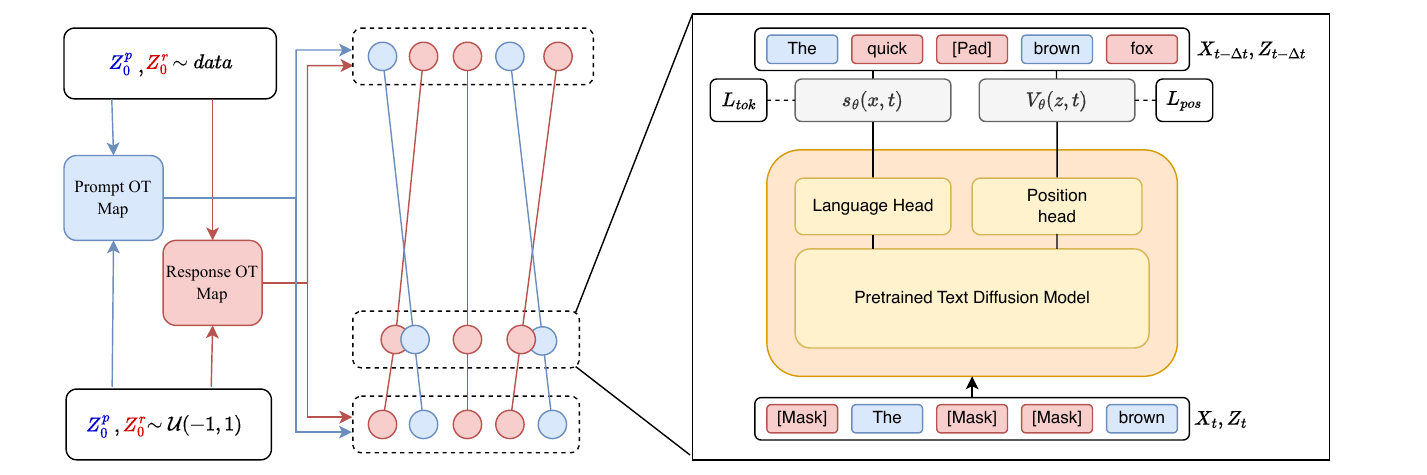}
    \caption{DDOT learns to vary infilled span lengths and positions, unlike prior fixed-position diffusion methods. \textit{(Left)} We compute two separate intra-set OT couplings within the prompt positions and the response positions, which drastically simplifies the set of possible permutations. \textit{(Right)} Given a time step $t$, we predict the token and position.}
    \label{fig:method}
\end{figure*}

\section{Related Work}

\subsection{Lexically Constrained Generation}
Constrained text generation has been explored through a variety of approaches, including AR and NAR methods \cite{zhang2020pointer, iso2022autotemplate, CBART, stern2019insertiontransformerflexiblesequence, InsNet}. POINTER \cite{zhang2020pointer} enables flexible token generation through iterative insertion, though it still depends on sequential token prediction and multiple forward passes. AutoTemplate \cite{iso2022autotemplate} approaches constrained text generation by feeding the prompt tokens into an encoder–decoder-style model. CBART \cite{CBART} extends the POINTER architecture by moving to an encoder–decoder model. Autoregressive methods, while effective for their specific use cases, inherit fundamental limitations: they require sequential generation that scales linearly with sequence length, and their causal attention masks prevent the full use of bidirectional context during generation. Most critically, for text infilling tasks, these approaches struggle to consider both past and future contexts simultaneously when generating intermediate content \cite{cao-etal-2023-improving}. Furthermore, to allow for flexible positions, autoregressive methods often regenerate the entire sequence instead of only inserting tokens, leading to wasted compute \cite{iso2022autotemplate}.

\subsection{Discrete Diffusion Models}
Discrete diffusion models offer an innovative approach to text generation, addressing key limitations of autoregressive methods \cite{lou2024discrete, ren2024discretecontinuousdiffusionmeet, gong2024scalingdiffusionlanguagemodels, sahoo2024simpleeffectivemaskeddiffusion}. These models denoise corrupted text sequences, enabling parallel updates of multiple tokens rather than the token-by-token process of autoregressive methods, thereby reducing the number of forward passes. Additionally, their bidirectional nature allows them to leverage both past and future context, in contrast to the causal masking that constrains autoregressive models. Early frameworks like D3PM \cite{austin2021structured} adapted continuous diffusion to discrete tokens by assigning probability that a token corrupts by either turning into a mask token or random token.

Recent work on score-based discrete diffusion has further advanced the field by providing analytical solutions for the denoising process. Instead of directly modeling transition probabilities, SEDD \cite{lou2024discrete} uses a score-based approach that learns the gradient of the log-probability and introduces an entropy-based loss function, narrowing the gap to autoregressive models.

However, despite these advantages, current discrete diffusion models face a significant limitation: they require fixed token positions throughout the generation process. This constraint makes them unsuitable for flexible-length text infilling, where the length of the generated text might differ from the original masked region.

\subsection{Optimal Transport Coupling for Flexible Generation}
OT \cite{villani2009optimal} coupling has been well studied in continuous diffusion. \citet{tong2024improving} introduce \emph{minibatch couplings}—either independent (random) or determined by OT—that regularize training and encourage near-straight trajectories. We adopt the term "coupling" from \citet{tong2024improving} to denote a matching between an source and target distribution. By incentivizing straighter paths, minibatch OT coupling yields practical benefits such as faster sampling, a pattern also observed in OT-inspired methods like Rectified Flow \citep{liu2023flow} and Stochastic Interpolants \citep{albergo2023building}.

Our OT coupling differs in scope and granularity: we construct a \emph{sample-level} coupling \emph{within each sequence}, aligning token positions between an initial state and a target ordering. We instantiate this coupling via OT (and compare to independent coupling in ablations), which preserves relative order and guides tokens along predictable paths while still allowing the inter-set crossings needed for flexible-length infilling. Unlike minibatch couplings aimed primarily at accelerating continuous diffusion models, our coupling enables joint optimization of \emph{content} and \emph{placement} of tokens for discrete text diffusion, retaining parallel and bidirectional conditioning and permitting dynamic adjustment of span lengths and token spacing.

\section{Background: Masked Text Diffusion}\label{sec:background}
Discrete diffusion adapts the continuous nature of diffusion processes to text. The simplest and most performant version of discrete diffusion is masked diffusion \cite{lou2024discrete, shi2024simplified, sahoo2024simpleeffectivemaskeddiffusion}. Rather than adding Gaussian noise to continuous values such as pixels, masked text diffusion assigns a probability of masking tokens throughout the forward diffusion process. For the purposes of this paper, masked diffusion can be seen as a masked language model (like BERT \cite{devlin2019bert}) that operates at progressively decreasing masking ratios to generate text. Specifically, our text diffusion process follows Score Entropy Discrete Diffusion (SEDD) \cite{lou2024discrete}, which models a score function over a support of $N$ states (token values). The forward diffusion process is described by a continuous-time Markov chain (CTMC):
\begin{equation}
    \frac{d p_{t}}{d t}=Q_{t} p_{t} \quad p_{0} \approx p_{\mathrm{data}},
    \label{eq:sedd-forward}
\end{equation}
where $p_t \in \mathbb{R}^{N}$ is the column vector of state probabilities at time $t \in [0,T]$, $Q_t \in \mathbb{R}^{N \times N}$ is the (time-dependent) transition matrix (its columns sum to $0$), and $p_{\mathrm{data}} \in \mathbb{R}^{N}$ is the empirical token distribution at $t=0$. In masked diffusion we include the mask token among the $N$ states and make it absorbing: each non-mask column has $-1$ on its diagonal and $+1$ in the mask row, while the mask column is all zeros, so columns sum to $0$ and probability flows from non-mask tokens into the mask token.

SEDD reverses the text diffusion dynamics by learning the score $s_{\theta}(x)_{y}$ of a transition from token $x$ to token $y$ via an entropy-based loss:
{\small
\begin{multline}
\mathcal{L}_{\text{tok}} = \int_0^T \mathbb{E}_{x_t \sim p_{t|0}(\cdot | x_0)} \sum_{y \neq x_t} Q_t(x_t, y)\,
\left( s_\theta(x_t, t)_y - \right. \\[-2pt]
\left. \frac{p_{t|0}(y|x_0)}{p_{t|0}(x_t|x_0)} \log s_\theta(x_t, t)_y
+ K\!\left(\frac{p_{t|0}(y|x_0)}{p_{t|0}(x_t|x_0)}\right) \right)\, dt
\label{eq:token-loss}
\end{multline}
}
where $x_0$ is the start token, $x_t$ is the token at time $t$ under the forward CTMC with marginal $p_{t|0}(\cdot|x_0)$, $Q_t(x_t,y)$ denotes the (row $x_t$, column $y$) entry of $Q_t$, and $K(a) = a (\log a - 1)$ is a scalar regularizer; Finally, to simulate the reverse diffusion process, we either take Euler steps or use an analytical denoiser \cite{lou2024discrete}.

\section{Approach}
Prior discrete diffusion models fix token positions during denoising and therefore require oracle positions to infill masked spans. We introduce \textbf{DDOT}, which jointly denoises discrete token values and continuous token positions so infilled spans can move and change length during generation. This enables flexible-length, flexible-position infilling from arbitrary prompts while preserving the parallel, bidirectional nature of discrete diffusion. \autoref{fig:method} summarizes DDOT.

\subsection{Token Value Diffusion}
We partition token values into three disjoint subsets \(x=(x^p, x^r, x^{\mathrm{pad}})\), which are described as follows:
(i) \emph{Prompt} (\(x^p\)): these slots are clamped to their ground-truth tokens for all \(t\); we never mask or resample them, and they serve purely as conditioning during denoising.
(ii) \emph{Response} (\(x^r\)): these are the real tokens to be infilled. They start at ground truth and follow the masked-diffusion forward process from \autoref{sec:background}, progressively transitioning toward \textsc{[mask]} as \(t\) increases; the reverse chain denoises them back to tokens.
(iii) \emph{Pad} (\(x^{\mathrm{pad}}\)): these slots hold the vocabulary token \textsc{[pad]} (distinct from \textsc{[mask]}), are not part of \(x^r\), and exist only to maintain a fixed length \(L\). Under the forward process (\autoref{sec:background}), \textsc{[pad]} may corrupt to \textsc{[mask]}; in the reverse process, \textsc{[mask]} on these slots denoises back to \textsc{[pad]}. 

\subsection{Continuous Position Diffusion}
To enable diffusion models to move masked tokens to regions that require infilling, DDOT also diffuses \emph{positions}. Similar to token values, we partition positions into three disjoint subvectors \(z_t=(z_t^p,\, z_t^r,\, z_t^{\mathrm{pad}})\) for prompt, response, and pad positions. Let \(\ell = |x^p|+|x^r|\) be the number of real tokens, with per-subset counts \(\ell^p=|x^p|\) and \(\ell^r=|x^r|=\ell-\ell^p\); there are \(L-\ell\) pad slots.

The noisy positions are sampled i.i.d.\ as \(z_T \sim \mathcal{U}(-1,1)^L\) (so \(z_T^p, z_T^r, z_T^{\mathrm{pad}}\) are slices of a single length-\(L\) vector).
We define the ground-truth positions for the \emph{real} (non-pad) tokens as the length-\(\ell\) subvector over the union of prompt and response indices,
\(z_0^{p\cup r} \in \mathbb{R}^{\ell}\) (equivalently \(z_0^{p\cup r}=(z_0^p, z_0^r)\)),
evenly spaced and scaled to the true length:
\begin{equation}
\label{eq:init_pos}
z^{p\cup r}_{0,i}
= \frac{\ell}{L}\,\frac{2i-(\ell-1)}{\max\{1,\ell-1\}},
\qquad i=0,\dots,\ell-1 .
\end{equation}
so real tokens occupy the sub-interval \([-\ell/L,\,\ell/L]\subset[-1,1]\).

\paragraph{Linear position paths.}
Following \citet{albergo2023building} and \citet{liu2023flow}, positions follow straight, noise-free interpolation:
\begin{equation}
\label{eq:linear-interp}
  z_t \;=\; (1-t)\,z_0 + t\,z_T .
\end{equation}

\paragraph{Pad tokens.}
Pad tokens are part of the ground-truth vocabulary (distinct from \textsc{[mask]}). Pad positions remain stationary up to the same global scaling:
\begin{equation}
\label{eq:pad-paths}
z^{\mathrm{pad}}_{0} \;=\; \frac{\ell}{L}\, z^{\mathrm{pad}}_{T}
\end{equation}
Together with the $z^{p\cup r}$ construction in \autoref{eq:init_pos}, this induces the full length-\(L\) slot vector \(z_0=(z_0^p, z_0^r, z_0^{\mathrm{pad}})\).

\subsection{Positional Optimal Transport Coupling}
Our preliminary experiments (\autoref{tab:OT-Ablation}) show that naively diffusing continuous token positions performs poorly. It induces a combinatorial explosion of permutations, weakens supervision from the ground truth, and leads to invalid infilling results. We therefore use \emph{sample-level} optimal-transport (OT) coupling. In particular, we find an OT coupling independently within each set $z^p$ and $z^r$. Our sample-level OT coupling has the following benefits:

\begin{description}[leftmargin=1.5em, labelsep=0.5em, itemsep=0pt, topsep=2pt, font=\normalfont\bfseries, style=standard]
  \item[(a) Reduces the permutation space.] OT yields order-preserving, non-crossing trajectories \emph{within each set} (prompt or response), drastically shrinking the search over permutations. Concretely, for either set \(S\in\{p,r\}\) and indices \(1\le i<j\le \ell^{S}\) in that set’s ground-truth order, we have \(z_t^{S,(i)} \le z_t^{S,(j)}\) for all \(t\in[0,1]\), so within-set crossings cannot occur (while inter-set prompt–response crossings remain allowed), avoiding the combinatorial explosion from arbitrary permutations.
  \item[(b) Preserves supervision signal.] Small word-order changes can flip semantics; The OT coupling between \(z_T\) and \(z_0\) preserves relative order throughout diffusion and thus maintains a stable learning signal from the ground-truth sequence rather than forcing the model to infer order from noisy, permuted trajectories.
  \item[(c) Prevents invalid permutations.] In infilling tasks, the prompt’s \emph{relative order is part of the conditioning}; any output that reorders prompt tokens is an invalid infilling result. We feed the prompt tokens at the terminal state \(T\) in their given order. With OT coupling, the forward \(0\!\to\!T\) paths are order-preserving within the prompt set (non-crossing), so the reverse diffusion returns the same within-set order at \(t=0\). Without coupling, trajectories can cross and swap prompt order, yielding invalid infills.
\end{description}

\paragraph{Assignment formulation (per set).}
We compute the optimal transport coupling \emph{within} prompt and non-prompt sets. From the noisy positions \(z_T\in\mathbb{R}^L\), sample \(\ell^p\) entries to form the prompt slice \(z_T^p\); the remaining \(L-\ell^p\) entries form the non-prompt slice \(z_T^{\mathrm{np}}\). We compare ground-truth positions (rescaled to \([-1,1]\)) directly to these slices.

\noindent\textbf{Prompt OT (balanced).}
One-to-one match: each of the \(\ell^p\) prompt positions at \(t=0\) is assigned to exactly one sampled prompt position at \(t=T\), minimizing total displacement:
\begin{equation}
\label{eq:prompt-ot}
\pi^p=\operatorname*{arg\,min}_{\pi\in S_{\ell^p}}
\sum_{i=1}^{\ell^p}\Bigl|\tfrac{L}{\ell}\,z^p_{0,i}-z^p_{T,\pi(i)}\Bigr|,
\quad
\tilde z_T^p = z_T^p[\pi^p].
\end{equation}
\noindent Where $S_{\ell^p}$ is the set of permutations of $[\ell^p]$ and, for any vector $v$,
$v[\pi]$ denotes reindexing by $\pi$: $(v[\pi])_i \coloneqq v_{\pi(i)}$.

\noindent\textbf{Response OT (unbalanced).}
Injective match: the \(\ell^r\) response positions at \(t=0\) are assigned into the non-prompt pool \(z_T^{\mathrm{np}}\) of size \(L-\ell^p\), each slot used at most once; unassigned slots become pads:
\begin{equation}
\label{eq:response-ot}
\pi^r=\operatorname*{arg\,min}_{\pi\in \Pi_{\ell^r}^{\,L-\ell^p}}
\sum_{i=1}^{\ell^r}\Bigl|\tfrac{L}{\ell}\,z^r_{0,i}-z^{\mathrm{np}}_{T,\pi(i)}\Bigr|,
\quad
\tilde z_T^r = z_T^{\mathrm{np}}[\pi^r].
\end{equation}
Here \(S_{\ell^p}\) is the set of all permutations of \([\ell^p]\), and
\(\Pi_{\ell^r}^{\,L-\ell^p}\) is the set of injective maps \([\ell^r]\hookrightarrow[L-\ell^p]\).

\noindent\textbf{Pads (leftover positions).}
The \(L-\ell\) non-prompt positions \emph{not} selected by \(\pi^r\) constitute \(z_T^{\mathrm{pad}}\). Pads are excluded from OT and follow the stationary path in \autoref{eq:pad-paths}.

After solving the assignments, we reorder the sampled positions and use the reordered slices for interpolation:
\begin{equation}
\label{eq:train-paths}
z_t^S \;=\; (1-t)\,z_0^S + t\,\tilde z_T^S,\qquad S\in\{p,r\} .
\end{equation}
This per-set OT coupling induces order-preserving, non-crossing trajectories within prompt and response (by construction of \(\tilde z_T^S\) and linear paths), while allowing \emph{inter-set} crossings between prompt and response. See \autoref{subsec:path_vis} for visualizations.

\paragraph{Computation.}
All couplings are computed \emph{per sequence} in one dimension, using absolute distance between rescaled ground-truth positions and sampled positions. This choice lets us avoid forming dense cost matrices that are typically used when calculating OT and enables simple, fast implementations that scale favorably in the sequence length.

\noindent\textbf{Prompt OT computation.}
For the prompt set we must match \(\ell^p\) ground-truth positions \(z_0^p\) to \(\ell^p\) sampled positions \(z_T^p\) in a one-to-one manner. In 1D with convex costs, the optimal assignment is equivalent to sorting both vectors and pairing by index. Therefore, we sort \(z_0^p\) and \(z_T^p\) once and obtain the permutation \(\pi^p\) that maps each sorted index in \(z_0^p\) to the corresponding rank in \(z_T^p\). This procedure achieves the exact minimum from \autoref{eq:prompt-ot}, runs in \(O(\ell^p\log \ell^p)\) time, and does not materialize a cost matrix.

\noindent\textbf{Response OT computation.}
We sort \(a=z_0^r\) and \(b=z_T^{\mathrm{np}}\) in ascending order, then compute the injective, order-preserving match with a standard global-alignment dynamic programming solution \citep{needleman1970general}. This exactly minimizes \autoref{eq:response-ot} and runs in \(O(\ell^r (L-\ell^p))\) time and \(O(L-\ell^p)\) memory.

\noindent\textbf{Complexity and practical overhead.}
The total per-sequence cost is \(O(\ell^p\log\ell^p + \ell^r(L-\ell^p))\) time and \(O(L)\) memory. For the context sizes considered (\(L\le 1024\)), the coupling time is negligible compared to a single denoiser forward pass. In practice we precompute several OT couplings in advance on the CPU and stream the dataset to the GPU, leading to negligible overhead, as seen in \autoref{subsec:analysis}


\subsection{Training Objective}
At each update we draw a time \(t\sim\mathcal{U}(0,1)\), sample noisy positions \(z_T\sim\mathcal{U}(-1,1)^L\), and form the matched targets \(\tilde z_T=(\tilde z_T^p,\tilde z_T^r,z_T^{\mathrm{pad}})\) by solving \autoref{eq:prompt-ot}–\autoref{eq:response-ot} (pads follow \autoref{eq:pad-paths}). We then construct the per-set paths (\autoref{eq:train-paths}). Independently, for token values we follow SEDD’s forward process (Sec.~\ref{sec:background}) to obtain \(x_t\) from \(x_0\) at the same \(t\).

\paragraph{Position loss.}
We supervise the position head (\autoref{fig:method}) on \emph{all} $L$ slots with a simple MSE toward the straight-path target:
\begin{equation}
\label{eq:Lpos}
\mathcal{L}_{\mathrm{pos}}(\theta)
\;=\;
\mathbb{E}_{t}\!\left[
\,\big\|\, v_\theta(z_t,x_t,t)\;-\; \bigl(z_0-\tilde z_T\bigr) \big\|_2^2
\right].
\end{equation}

\paragraph{Token loss.}
For tokens we use the SEDD score-entropy objective from \autoref{sec:background}, denoted \(\mathcal{L}_{\mathrm{tok}}(\theta)\).

\paragraph{Total loss.}
The final objective is a weighted sum
\begin{equation}
\label{eq:total-loss}
\mathcal{L}(\theta)
\;=\; 
\mathcal{L}_{\mathrm{tok}}(\theta)
\;+\;
\lambda\,\mathcal{L}_{\mathrm{pos}}(\theta),
\end{equation}
with \(\lambda\) controlling the relative weight of position supervision.

\paragraph{Initialization: DDOT-R vs.\ DDOT-U.}
During inference we manipulate how we sample $z_T$. By default (DDOT-R), we randomly sample a full length-$L$ terminal vector $z_T\sim\mathcal{U}(-1,1)^L$, reserve any $\ell^p$ entries as the prompt slice $z_T^p$. However, random sampling can produce high-density regions that tend to map to pad tokens because there are fewer tokens in the corresponding ground-truth region.

To mitigate this, DDOT-U places terminal positions on uniform grids for each set:
\begin{equation}
\label{eq:term_pos}
\begin{aligned}
z^{p}_{T,i} \;=\; \frac{\,2i-(\ell^{p}-1)\,}{\max\{1,\;\ell^{p}-1\}}, \qquad & i=0,\dots,\ell^{p}-1,\\
z^{np}_{T,i} \;=\; \frac{\,2i-(\ell^{np}-1)\,}{\max\{1,\;\ell^{np}-1\}}, \qquad & i=0,\dots,\ell^{np}-1.
\end{aligned}
\end{equation}
Uniform spacing avoids random clustering, reduces pad spillover, and yields more stable placements.

\paragraph{Reverse-time updates.}
During sampling, positions follow near-straight trajectories because the learned velocity field approximates the constant ground-truth direction $z_0 - \tilde{z_T}$. We therefore update positions with a simple Euler step at each $\tau$-leap of the SEDD reverse process, while keeping prompts clamped and pads stationary (\autoref{eq:pad-paths}). 

\subsection{Simultaneous Text \& Position Diffusion}
DDOT performs discrete text and continuous position diffusion simultaneously, as these processes operate independently in continuous time. We therefore predict both token-value scores and position velocities in a single forward pass. This independence also enables simulation-free training by sampling token and position states independently at arbitrary time steps. We summarize the training procedure in algorithm \autoref{alg:ddot}.

\begin{algorithm}[!hbtp]
\caption{DDOT Training}
\label{alg:ddot}
\small
\begin{algorithmic}[1]
\Require batch $(x_0, z_0)$; loss weight $\lambda$
\State Sample $t \sim \mathcal{U}(0,1)$
\State Build terminal positions: sample $z_T \sim \mathcal{U}(-1,1)^L$ and split $(z_T^p, z_T^{\mathrm{np}})$, or use uniform grids from \autoref{eq:term_pos}
\State Compute within-set OT: $\pi^p$ via 1D sort-match (\autoref{eq:prompt-ot}); $\pi^r$ via injective matching (\autoref{eq:response-ot}); set $\tilde z_T^p=z_T^p[\pi^p]$, $\tilde z_T^r=z_T^{\mathrm{np}}[\pi^r]$, leftovers $\to z_T^{\mathrm{pad}}$
\State Form paths $z_t^S=(1-t)z_0^S + t\,\tilde z_T^S$ for $S\in\{p,r\}$; set pads via \autoref{eq:pad-paths}
\State Sample token states $x_t$ from the SEDD forward at time $t$
\State Predict $s_\theta(x_t,t)$ and $v_\theta(z_t,x_t,t)$
\State Compute $\mathcal{L}_{\mathrm{tok}}$ (\autoref{eq:token-loss}) and $\mathcal{L}_{\mathrm{pos}}$ (\autoref{eq:Lpos})
\State Update $\theta$ on $\mathcal{L}_{\mathrm{tok}} + \lambda\,\mathcal{L}_{\mathrm{pos}}$ (\autoref{eq:total-loss})
\end{algorithmic}
\end{algorithm}
\subsection{Implementation Details}
We extend SEDD, which is based on the Diffusion Transformer architecture \cite{peebles2023scalable}, with two additional modules. First, we introduce a learnable type embedding applied directly after the token-embedding lookup. This embedding indicates whether a token is part of the prompt or the masked response (\(x_i \in x^p\) or \(x_i \in x^np\)), which is critical for assigning each token to the correct OT flow. Second, we add a linear head at the end of the Diffusion Transformer to compute \(v_\theta(z_t, t)\). The model architecture is available in \autoref{fig:method}.

To incorporate continuous positional information, we scale \(z_t\) from \([-1, 1]\) to match the context length of the original pretrained model (1024). We then use Rotary Position Embeddings \cite{su2024roformer}, a standard technique in discrete diffusion models. Implementation details can be found in Subsection~\ref{subsec:hyperparam}.

\section{Experiments}

\subsection{Experimental Setup}
\noindent\textbf{Datasets}\hspace{0.75em}
We evaluate our approach on the One-Billion-Word and Yelp datasets, following the preprocessing steps outlined in prior work on infilling and lexically constrained generation \cite{miao2019cgmh, zhang2020pointer, iso2022autotemplate}. These datasets consist of examples with $1$ to $6$ keywords that must be infilled while maintaining their relative order to generate coherent sentences. In addition to randomly masking positions, we also introduce a \emph{block} masking method that masks a single continuous span of text ranging from $0$ to $L/2$ tokens (32 for One-Billion-Word and Yelp; 512 for CodeParrot). Finally, we apply the aforementioned masking methods to the Python subset of the CodeParrot dataset. \autoref{tab:example-generations} illustrates examples of this lexically constrained generation task.

\input{tables/table2}
\input{tables/table5.tex}
\input{tables/table1.tex}

\noindent\textbf{Training Details}\hspace{0.75em}
To align the position-prediction modules, we first fine-tune SEDD with the added modules on FineWeb-Edu \cite{penedo2024fineweb}. Afterward, we further fine-tune on the One-Billion-Word and Yelp datasets. For simplicity, we keep all parameters unfrozen and optimize $\mathcal{L}_{\text{tok}}$ and $\mathcal{L}_{\text{pos}}$ simultaneously.

In line with SEDD, we train our model in two configurations: \emph{small} (90M non-embedding parameters) and \emph{medium} (320M non-embedding parameters). DDOT-medium is on the same scale as CBART (406M parameters) and AutoTemplate-base (220M parameters). Following SEDD, we use the AdamW optimizer with a learning rate of \(3\times 10^{-5}\). We set $\lambda=10$ for position-loss weighting. For each experiment, we use either $48\times$ L40 (48\,GB), $80\times$ A30 (24\,GB), or $8\times$ A100 (80\,GB) GPUs.\\

\noindent\textbf{Baselines}\hspace{0.75em}
We compare our method against strong autoregressive (AR) and non-autoregressive (NAR) baselines. AutoTemplate \cite{iso2022autotemplate}, a state-of-the-art AR approach, leverages the T5 \cite{raffel2020exploring} family of pretrained models and parses the lexically constrained generation task into a template that is generated autoregressively from left to right. The previous state-of-the-art NAR method, CBART \cite{CBART}, is built on the pretrained BART framework \cite{lewis2020bart} and iteratively inserts tokens into a sequence.

We also introduce two diffusion-based baselines that follow the same training procedure as DDOT. \emph{Left Context (LC)} concatenates all prompt tokens to the left of the sequence and generates the response to the right of a separator token. \emph{Position Prediction (PoP)} uses a SEDD model with a linear head that first predicts the positions of every token; this sequence is then fed through a fine-tuned fixed-position SEDD.\\
  
\noindent\textbf{Distribution Annealing}\hspace{0.75em}
Many lexically constrained generation baselines, including AutoTemplate and CBART, use distribution-annealing methods such as top-$p$, top-$k$, greedy decoding, and beam search. To provide a parallel to greedy decoding—which always takes the top token probability—we anneal the token-value distribution during sampling to include only the most probable non-mask token. Specifically, given the predicted probability that a token is the mask, $\hat{p}(x^{\text{mask}})$, we assign $1 - \hat{p}(x^{\text{mask}})$ to the token value with the highest probability excluding the mask token, and set the remaining token probabilities to $0$. Greedy decoding in prior models (e.g., AR) is deterministic, collapsing the tree of all generation paths into a single path. However, our annealing process maintains generation diversity (\ref{subsec:gen_div}) because the model still samples from the annealed distribution over the top token value and the mask token. Whenever possible, we evaluate against the greedy-decoding baseline. \\

\noindent\textbf{Metrics}\hspace{0.25em}
Following prior work \cite{miao2019cgmh, zhang2020pointer, iso2022autotemplate}, we evaluate BLEU-2/4 \cite{papineni2002bleu}, NIST-2/4 \cite{doddington2002automatic}, METEOR-v1.5 \cite{denkowski2014meteor}, and success rate (SR), which is the percentage of responses that are valid infilling results. A result is determined as invalid when any prompt tokens in the generated sequence do not appear in the same relative order that they were provided.

\subsection{Main Results}
We present lexically constrained generation results in \autoref{tab:results_ours}. Our approach uses greedy annealing and is compared against greedy decoding wherever applicable, including the CBART greedy-decoding baseline. Our method achieves competitive performance with previous NAR models, approaching AR performance. Notably, our model achieves state-of-the-art performance on most metrics among the diffusion baselines. Our method performs well on block infilling, which may be more useful in real-world applications. Furthermore, we observe that DDOT scales well to longer sequences: it frequently generates valid responses that include all prompt words in the same relative order, as reflected in the SR. In contrast, \textbf{diffusion baselines quickly generate invalid responses} as the number of prompt tokens increases (\autoref{fig:num_prompt}, \autoref{tab:results_ours}). However, in benchmarks with six or fewer prompt tokens, diffusion baselines maintain high SR (\autoref{tab:results}). This may be because fixed-position models have room to correct generation when the ratio of prompt to response tokens is low. 

\autoref{tab:results} compares results with previous work in lexically constrained generation. Since pretrained diffusion models currently lag behind AR models—an issue not unique to DDOT—we focus on NAR models. DDOT performs on par with previous state-of-the-art models and achieves higher SR than all diffusion baselines. Although DDOT underperforms LC and PoP on some metrics, we argue that the One-Billion-Word–Random and Yelp–Random settings over-index on the unrealistic task of generating text from only $1$–$6$ randomly spaced tokens. Furthermore, \autoref{fig:num_prompt} shows the broader trend when DDOT is scaled to a larger number of prompt tokens.

\subsection{Analysis}
\label{subsec:analysis}
In this section, we investigate the effect of random versus uniform position initialization, the inclusion of OT coupling, and the impact of varying the number of sampling steps. \\

\begin{figure*}[!th]
  \centering
    \includegraphics[width=\textwidth]{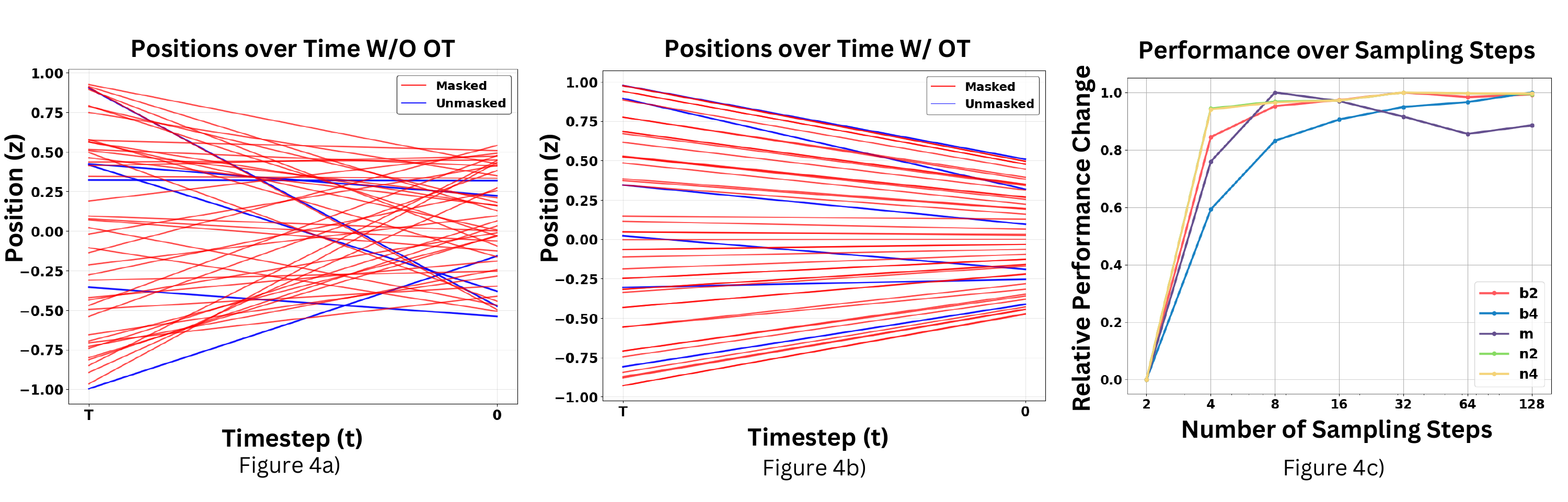}
    \vspace{-2.5em}
    \caption{(a) and (b) show ground-truth token positions over time. Without OT (a), many line crossings indicate unstable permutations, whereas with OT coupling (b), trajectories are nearly straight throughout the denoising process. (c) Performance tends to increase with more sampling steps.}
    \vspace{-1em}
  \label{fig:OT_NOT-visualization}
\end{figure*}

\noindent\textbf{Position Initialization}\hspace{0.75em}
In \autoref{tab:results_ours} and \autoref{tab:results}, we also explore the difference between DDOT-R and DDOT-U. In DDOT-R, pad tokens tend to cluster in high-density regions because OT finds no matches for them; in DDOT-U, pad tokens tend to be evenly spaced. We find that DDOT-U consistently outperforms DDOT-R.\\

\noindent\textbf{OT Coupling}\hspace{0.75em}
To demonstrate the importance of OT coupling between source and target positions, we retrain the small version of DDOT without OT coupling and provide a quantitative comparison in \autoref{tab:OT-Ablation}. Models trained with OT coupling consistently outperform those using independent (random) coupling. We hypothesize that OT coupling provides a stronger signal about token ordering throughout the diffusion process. Specifically, DDOT guarantees that the relative ordering of the prompt and generated tokens at any time step matches the original ordering. In contrast, independent coupling requires the model to infer the original ordering—a challenging task given the numerous plausible orderings that can result from interspersed prompt tokens.
\input{tables/table3.tex}

\noindent\textbf{Position Over Time}\hspace{0.75em}
We qualitatively compare ground-truth token paths during training in \autoref{fig:OT_NOT-visualization}. With OT coupling, token trajectories exhibit significantly fewer crossings, maintaining relative order throughout the generation process; in contrast, independent coupling frequently permutes tokens. Visualizations of token paths during inference are available in \autoref{subsec:path_vis}.\\

\noindent\textbf{Number of Sampling Steps}\hspace{0.75em}
One advantage of diffusion models over autoregressive models is the ability to trade compute for accuracy by varying the number of inference steps. \autoref{fig:OT_NOT-visualization} shows how the number of sampling steps influences lexically constrained generation performance: as the number of sampling steps increases, performance improves.\\

\noindent\textbf{Wall-Time Analysis}\hspace{0.75em}
We evaluate the inference speed of DDOT against the diffusion baselines on the One-Billion-Word. \autoref{tab:inference_speed} presents the wall-clock time per batch alongside BLEU-2 and BLEU-4 scores for increasing numbers of sampling steps.

DDOT demonstrates significantly better efficiency. For any given number of sampling steps, DDOT is not only faster than LC and competitive with PoP in raw speed, but also achieves substantially higher BLEU scores. Notably, LC must regenerate prompt tokens and therefore requires up to double the input sequence length. PoP also requires an additional forward pass to predict initial positions.\\

\noindent\textbf{Efficiency Considerations}\hspace{0.75em}
\input{tables/table_new}
The added modules that enable position prediction are lightweight, consisting of a linear head and two type embeddings. By contrast, the LC baseline requires double the context length of DDOT because it must regenerate prompt tokens.

The OT calculation is highly efficient, taking 16 minutes, 11 seconds on an Intel Xeon 8462Y+ 64-core processor for the 10-billion-token subset of FineWeb-Edu. In practice, we stream the dataset, caching several OT couplings in advance without needing to preprocess them. With caching, it takes 4 minutes, 30 seconds to run 1{,}000 training steps on an L40 GPU with a batch size of 256. Without caching, it takes 4 minutes, 27 seconds—a negligible difference.

\section{Conclusion}
In this work, we introduce DDOT, the first discrete diffusion model capable of flexible-length text infilling by jointly denoising token values and positions. By incorporating optimal-transport coupling, DDOT preserves token order while enabling dynamic position adjustment, addressing limitations in existing text diffusion models. Our experiments show that DDOT outperforms diffusion baselines and matches state-of-the-art non-autoregressive models.

\section*{Limitations}
DDOT inherits many drawbacks from previous discrete text diffusion methods \cite{lou2024discrete, austin2021structured, sahoo2024simpleeffectivemaskeddiffusion}. First, diffusion backbones still lag behind autoregressive backbones. Since diffusion for text is an emerging field, this performance gap is expected. Similarly, the fine-tuned versions of these models (e.g., DDOT vs.\ AutoTemplate) exhibit the same gap. Second, unlike autoregressive models, existing diffusion models cannot use a KV cache to store previous activations. DDOT also inherits risks common to large-scale text models.

\section*{Ethical Considerations}
\noindent\textbf{Artifacts}\quad The artifacts we used have public-use licenses. Furthermore, we plan to release our code artifacts for public use after acceptance.\\

\noindent\textbf{Dataset Considerations}\quad We use publicly available datasets that underwent safety checks, such as FineWeb-Edu.\\

\noindent\textbf{Documentation of Artifacts}\quad Our work generates English text, and our codebase is primarily in Python.\\

\noindent\textbf{Use of AI Assistants}\quad We used AI assistants to help write our code and revise our paper.\\

\noindent\textbf{Package Usage}\quad We use NLTK for BLEU, NIST, METEOR, and $n$-gram diversity. We use \url{https://pypi.org/project/codebleu/} for CodeBLEU.

\section*{Acknowledgements}
We thank Virginia Tech’s Advanced Research Computing (ARC) for computational resources and technical support, and the anonymous reviewers for their constructive feedback.

\bibliography{custom}

\clearpage
\onecolumn
\appendix

\clearpage

\begingroup
\setlength{\textfloatsep}{8pt}
\setlength{\floatsep}{8pt}
\setlength{\intextsep}{8pt}

\section{Appendix}
\label{sec:appendix}

\subsection{Overview}
This appendix provides additional results, diversity analyses, hyperparameter sensitivity, path visualizations, and short implementation notes for reproducibility. We reuse notation from the main text and make one explicit addition used below:
\[
\ell^{\mathrm{np}} \;=\; L - \ell^{p}
\]
denoting the number of non-prompt slots (response \(+\) pads). This matches the usage of \(z_T^{\mathrm{np}}\) in the main text.

\subsection{Additional Results}
\label{app:addl_results}

\subsubsection{Size Comparison}
\label{subsec:size_comp}
Table~\ref{tab:results_DDOT-small} reports BLEU-2/4, NIST-2/4, and METEOR for \textsc{ddot} small/medium and for \textsc{ddot-R} vs.\ \textsc{ddot-U} on One-Billion-Word–Random and Yelp–Random. Uniform \(z_T\) initialization (\textsc{ddot-U}) consistently outperforms random (\textsc{ddot-R}), and scaling to the medium model further improves most metrics.
\input{tables/table6.tex}

\subsubsection{Generation Diversity}
\label{subsec:gen_div}
Table~\ref{tab:results_diversity} compares diversity using D2/D4 (unique bigrams/four-grams). \textsc{ddot-U} yields competitive diversity relative to strong NAR baselines while maintaining ordering constraints.
\input{tables/table7}

\subsection{Hyperparameter Sensitivity and Training Details}
\label{app:hyper}

\subsubsection{Loss Weight \(\lambda\)}
\label{app:lambda}
We sweep \(\lambda\) to balance \(\mathcal{L}_{\text{tok}}\) and \(\mathcal{L}_{\text{pos}}\). Results in Table~\ref{tab:lambda_values} show stable performance across a range; \(\lambda\!\approx\!10\) is a good default used in the main results.
\input{tables/table8}

\subsubsection{Training Hyperparameters}
\label{subsec:hyperparam}
Table~\ref{tab:train_hyperparams} lists optimizer, learning rate, weight decay, \(\beta\)s, batch sizes (by GPU), and schedules. Unless noted, we follow SEDD defaults and fine-tune all parameters jointly.
\input{tables/table9}

\subsection{Path Visualizations}
\label{subsec:path_vis}

We visualize token-position \emph{trajectories} during both \textbf{inference} (what the model actually does at test time) and \textbf{training} (the straight paths used as supervision). We show two coordinate systems: \textbf{scaled} to $[-1,1]$ (used in losses; cf.\ \autoref{eq:init_pos}) and \textbf{unscaled} in the model’s native context length (for intuition about absolute movement). Unless noted, we split by set $S\!\in\!\{p,r\}$ and index tokens within a set by their ground-truth order.

\paragraph{Plot Explanation.}
Let $\{t_u\}_{u=0}^{K}$ be the reverse-time grid (left-to-right on the $x$-axis), with $t_0=T$ and $t_K=0$. For each token index $i$ in set $S$, we plot its 1D position $z^{S,(i)}_{t_u}$ across $u$. Colors denote the token’s state (masked vs.\ unmasked) at each step. During inference, $z_{t_{u+1}}$ is obtained by an Euler update using the learned velocity $v_\theta(z_{t_u},x_{t_u},t_u)$ while clamping prompts and keeping pads stationary (\autoref{eq:pad-paths}). During training, targets follow straight paths $z_t^S=(1-t)z_0^S+t\,\tilde z_T^S$ from \autoref{eq:train-paths}.

\paragraph{Analysis.}
Within a \emph{set} (prompt or response), desirable behavior is (i) \textbf{order preservation} (no within-set crossings) and (ii) \textbf{near-straight trajectories} (little curvature). \emph{Inter-set} crossings (prompt vs.\ response) are allowed—and often required—to place response tokens between prompts during infilling.

\begin{figure*}[t]
  \centering
  \begin{subfigure}{0.85\textwidth}
    \centering
    \includegraphics[width=\linewidth]{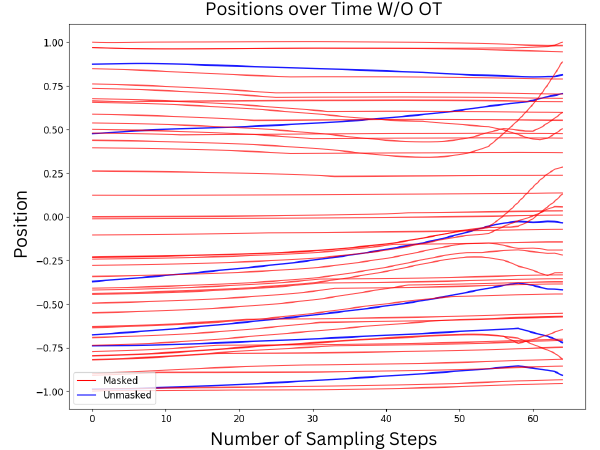}
    \caption{Scaled positions over time without OT.}
    \label{fig:not_scaled_appendix}
  \end{subfigure}

  \vspace{0.6em}

  \begin{subfigure}{0.85\textwidth}
    \centering
    \includegraphics[width=\linewidth]{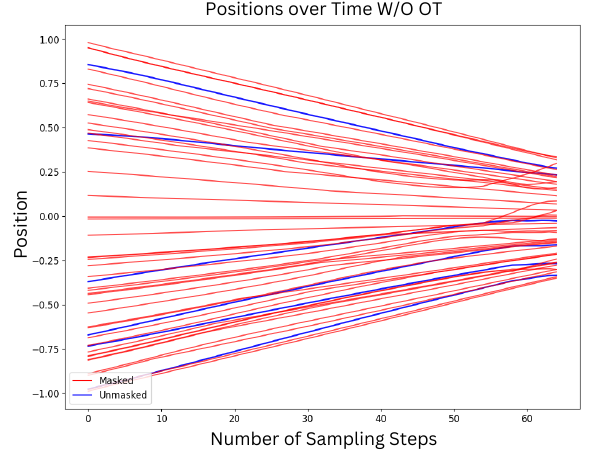}
    \caption{Unscaled positions over time without OT.}
    \label{fig:not_unscaled_appendix}
  \end{subfigure}
  \caption{Ablations on number of sampling steps without OT coupling. Paths curve and cross more often, complicating learning and inference.}
  \label{fig:NOT-visualization_appendix}
\end{figure*}

\begin{figure*}[t]
  \centering
  \begin{subfigure}{0.85\textwidth}
    \centering
    \includegraphics[width=\linewidth]{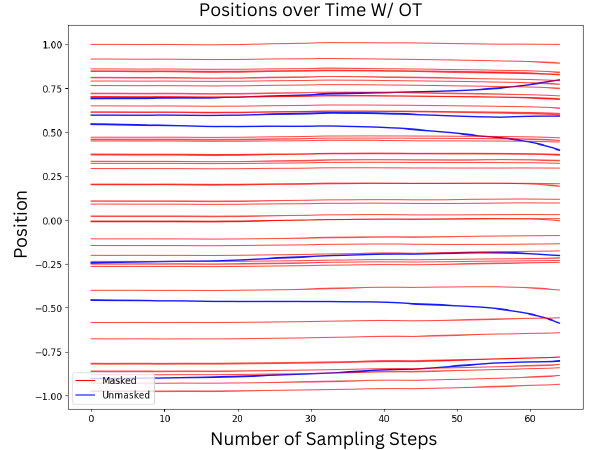}
    \caption{Scaled positions over time with OT.}
    \label{fig:ot_scaled_appendix}
  \end{subfigure}

  \vspace{0.6em}

  \begin{subfigure}{0.85\textwidth}
    \centering
    \includegraphics[width=\linewidth]{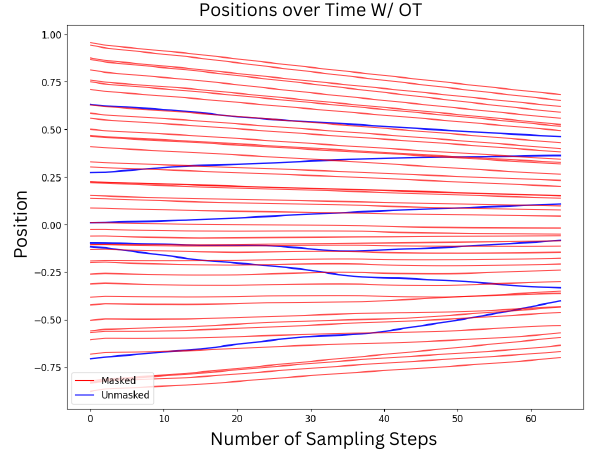}
    \caption{Unscaled positions over time with OT.}
    \label{fig:ot_unscaled_appendix}
  \end{subfigure}
  \caption{With per-set OT coupling, trajectories are straighter and order-preserving within each set.}
  \label{fig:OT-visualization_appendix}
\end{figure*}

\begin{figure*}[t]
  \centering
  \begin{subfigure}{0.85\textwidth}
    \centering
    \includegraphics[width=\linewidth]{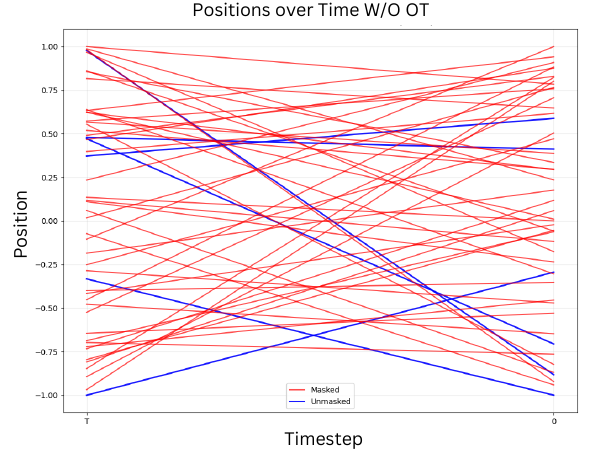}
    \caption{Scaled training targets without OT: many crossings.}
    \label{fig:no_ot_gt_appendix}
  \end{subfigure}

  \vspace{0.6em}

  \begin{subfigure}{0.85\textwidth}
    \centering
    \includegraphics[width=\linewidth]{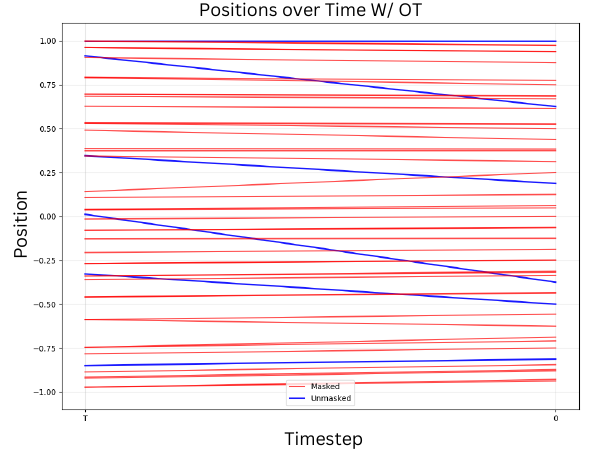}
    \caption{Scaled training targets with OT: straighter, non-crossing (within-set).}
    \label{fig:ot_gt_appendix}
  \end{subfigure}

  \caption{Ground-truth token paths (scaled to $[-1,1]$) used as supervision. Without OT (left), many crossings indicate unstable matching; with OT (right), trajectories are nearly straight and order-preserving within each set.}
  \label{fig:OT_NOT_gt_velocities_appendix}
\end{figure*}

\subsection{Implementation Notes}
\label{app:impl}
\textbf{Computing OT efficiently.} Prompt matching is solved exactly in 1D by sorting \(z_0^p\) and \(z_T^p\) and pairing by rank; response matching uses order-preserving injective alignment via global-alignment DP against \(z_T^{\mathrm{np}}\). Both avoid dense cost matrices and run in \(O(\ell^p\log\ell^p + \ell^r(L-\ell^p))\) time and \(O(L)\) memory.

\textbf{Caching \& streaming.} For large corpora, precompute couplings on CPU and stream mini-batches; we observed negligible overhead whether caching or computing couplings on the fly.

\endgroup
\end{document}

%% file: tables/table2.tex
\begin{table}[H]
    \centering
    \small
    \begin{tabular}{p{2cm}p{5cm}}
        \toprule
        Keywords: & \colorbox{pink}{earned}, \colorbox{blue!20}{cents}, \colorbox{yellow!50}{share}, \colorbox{green!30}{costs} \\
        \midrule
        One-Billion-Word-Random: & It \colorbox{pink}{earned} 28 \colorbox{blue!20}{cents} a \colorbox{yellow!50}{share} , excluding restructuring \colorbox{green!30}{costs}. \\
        \midrule \midrule
        Keywords: & \colorbox{pink}{we are so incredibly wedding .}\\
        \midrule 
        Yelp-Block: & \colorbox{pink}{we are so incredibly} happy we chose this venue for our \colorbox{pink}{wedding .} \\
        \bottomrule
    \end{tabular}
    \caption{Example generations for the keywords-to-sentence generation on One-Billion-Word and Yelp.}
    \vspace{-1.5em}
    \label{tab:example-generations}
\end{table}

%% file: tables/table5.tex
\begin{table*}[ht]
\centering
\resizebox{\textwidth}{!}{%
\begin{tabular}{l|cccc|cccc|ccc|ccc}
\toprule
\textbf{Model} & \multicolumn{4}{c|}{\textbf{One Billion Word--Block}} & \multicolumn{4}{c|}{\textbf{Yelp-Block}} & \multicolumn{3}{c|}{\textbf{CodeParrot-Block}} & \multicolumn{3}{c}{\textbf{CodeParrot-Random}} \\
               & B2/B4   & N2/N4   & M    & SR & B2/B4   & N2/N4   & M  & SR & B2/B4   & CB   & SR   & B2/B4   & CB   & SR   \\ \midrule
LC                     & 55.1/47.9  & 4.38/4.40 & 37.4 & 54.7 & 59.1/51.7  & 7.77/8.48 & 37.9 & 60.3 & 51.0/49.4 & 53.3 & \textbf{16.9} & 15.4/24.8 & 19.4 & 0. \\
PoP                    & 59.2/50.0  & \textbf{8.43}/8.77 & 37.1 & 58.6 & 64.9/48.6 & 7.81/8.47 & 38.1 & 67.1 & 29.4/13.3 & 21.7 & 0.34 & 12.5/2.75 & 13.9 & 0. \\
DDOT-R                  & 62.2/57.1  & 7.42/8.16 & 42.1 & 99.7  & 62.5/57.3 & 7.50/8.24 & 42.2 & 99.8 & 47.2/41.1 & 38.5 & 1.01 & 57.2/43.5 & 39.2 & 3.53   \\ 
DDOT-U                     & \textbf{63.7}/\textbf{58.4}  & 8.40/\textbf{8.79} & \textbf{42.1} & \textbf{100.} & \textbf{64.9}/\textbf{59.5} & \textbf{8.06}/\textbf{8.86} & \textbf{42.9} & \textbf{100.} & \textbf{53.7}/\textbf{51.0} & \textbf{50.5} & 10.8 & \textbf{58.2}/40.4 & \textbf{45.4} & \textbf{11.2}  \\

\bottomrule
\end{tabular}%
}
\caption{\textbf{DDOT outperforms diffusion baselines on standard sequences (0-32 prompt tokens).} Metrics are BLEU (B2, B4), NIST (N2, N4), METEOR (M), and Success Rate (SR). Top scores are \textbf{bolded}.}
\label{tab:results_ours}
\end{table*}

%% file: tables/table1.tex
\begin{table*}[ht]
\begin{minipage}{0.65\textwidth}
\raggedright
\resizebox{\textwidth}{!}{%
\begin{tabular}{p{5.5cm}|cccc|cccc}
\toprule
\textbf{Model} &
  \multicolumn{4}{c|}{\textbf{One Billion Word--Random}} &
  \multicolumn{4}{c}{\textbf{Yelp-Random}}\\
& B2/B4 & N2/N4 & M & SR & B2/B4 & N2/N4 & M & SR\\
\midrule
\multicolumn{9}{l}{\textcolor{gray}{\textbf{Autoregressive Models}}}\\
\midrule
\textcolor{gray}{GBS} \textcolor{gray}{\cite{hokamp2017lexically}} &
 \textcolor{gray}{10.1}/\textcolor{gray}{2.8} &
 \textcolor{gray}{1.49}/\textcolor{gray}{1.50} &
 \textcolor{gray}{13.5} &
 \textcolor{gray}{$\leq 100.$} &
 \textcolor{gray}{13.6}/\textcolor{gray}{4.5} &
 \textcolor{gray}{1.68}/\textcolor{gray}{1.71} &
 \textcolor{gray}{15.3} &
 \textcolor{gray}{$\leq 100.$}\\

\textcolor{gray}{InstructGPT} \textcolor{gray}{\cite{ouyang2022training}} &
 \textcolor{gray}{10.1}/\textcolor{gray}{2.8} &
 \textcolor{gray}{1.72}/\textcolor{gray}{1.73} &
 \textcolor{gray}{13.0} &
 \textcolor{gray}{92.33} &
 \textcolor{gray}{9.3}/\textcolor{gray}{2.4} &
 \textcolor{gray}{1.42}/\textcolor{gray}{1.44} &
 \textcolor{gray}{13.6} &
 \textcolor{gray}{92.17}\\

\textcolor{gray}{AutoTemplate-small} \textcolor{gray}{\cite{iso2022autotemplate}} &
 \textcolor{gray}{16.4}/\textcolor{gray}{6.1} &
 \textcolor{gray}{3.11}/\textcolor{gray}{3.15} &
 \textcolor{gray}{15.5} &
 \textcolor{gray}{100.} &
 \textcolor{gray}{22.5}/\textcolor{gray}{9.5} &
 \textcolor{gray}{3.51}/\textcolor{gray}{3.63} &
 \textcolor{gray}{17.1} &
 \textcolor{gray}{100..}\\

\textcolor{gray}{AutoTemplate-base} \textcolor{gray}{\cite{iso2022autotemplate}} &
 \textcolor{gray}{18.3}/\textcolor{gray}{7.6} &
 \textcolor{gray}{3.39}/\textcolor{gray}{3.45} &
 \textcolor{gray}{16.0} &
 \textcolor{gray}{100.} &
 \textcolor{gray}{23.7}/\textcolor{gray}{10.8} &
 \textcolor{gray}{3.62}/\textcolor{gray}{3.76} &
 \textcolor{gray}{17.8} &
 \textcolor{gray}{100..}\\

\textcolor{gray}{AutoTemplate-large} \textcolor{gray}{\cite{iso2022autotemplate}} &
 \textcolor{gray}{18.9}/\textcolor{gray}{8.1} &
 \textcolor{gray}{3.49}/\textcolor{gray}{3.54} &
 \textcolor{gray}{16.2} &
 \textcolor{gray}{100.} &
 \textcolor{gray}{24.1}/\textcolor{gray}{11.1} &
 \textcolor{gray}{3.68}/\textcolor{gray}{3.83} &
 \textcolor{gray}{17.9} &
 \textcolor{gray}{100..}\\
\midrule
\multicolumn{9}{l}{\textbf{Non-Autoregressive Models}}\\
\midrule
\multicolumn{9}{l}{\textit{Traditional Models}}\\
\hspace{1em}SeqBF \cite{mou2016sequence} &
 4.4/0.7 & 0.62/0.62 & 7.0 & $\leq 100.$ &
 6.9/2.1 & 0.52/0.53 & 8.7 & $<100.$\\
\hspace{1em}CGMH \cite{miao2019cgmh} &
 9.9/3.5 & 1.15/1.17 & 13.1 & \textbf{100.} &
 12.3/4.6 & 1.41/1.45 & 14.6 & $\leq 100.$\\
\hspace{1em}POINTER \cite{zhang2020pointer} &
 8.7/1.6 & 2.11/2.12 & 14.3 & \textbf{100.} &
 10.6/2.4 & 2.14/2.16 & 16.8 & \textbf{100.}\\
\hspace{1em}CBART \cite{CBART} &
 15.6/\textbf{6.6} & 2.16/2.19 & \textbf{15.2} & \textbf{100.} &
 19.4/\underline{9.0} & 2.54/2.64 & 17.4 & \textbf{100.}\\
\midrule
\multicolumn{9}{l}{\textit{Diffusion Models}}\\
\hspace{1em}LC &
 15.36/\underline{6.52} & 2.02/2.05 & 14.99 & \underline{99.82} &
 20.55/\textbf{9.66} & 2.76/2.87 & 17.58 & 99.75\\
\hspace{1em}PoP &
 \textbf{16.59}/5.66 & 3.06/3.09 & \underline{15.15} & 99.05 &
 \underline{21.13}/7.97 & \underline{3.31}/\underline{3.40} &
 \textbf{17.90} & 98.87\\
\hspace{1em}DDOT-R (Ours) &
 15.7/5.1 & \textbf{3.17}/\textbf{3.21} & 14.6 & 99.60 &
 19.7/7.0 & 3.26/3.34 & 17.4 & \underline{99.77}\\
\hspace{1em}DDOT-U (Ours) &
 \underline{16.3}/5.2 & \underline{3.13}/\underline{3.17} & 15.0 & \textbf{100.} &
 \textbf{21.2}/7.9 & \textbf{3.43}/\textbf{3.52} &
 \underline{17.7} & \textbf{100.}\\
\bottomrule
\end{tabular}}
\setlength{\abovecaptionskip}{1.2em}
\caption{\textbf{DDOT performs on-par with state-of-the-art NAR models on short sequences (1–6 prompt tokens).}
Top NAR scores are \textbf{bold}; second-best are \underline{underlined}.
Since SOTA NAR backbones (e.g.\ diffusion) still lag behind AR backbones,
we focus on NAR comparisons.}
\label{tab:results}
\end{minipage}%
\hspace{1em}%
\begin{minipage}{0.33\textwidth}
\raggedleft
\vspace{-1.2em}
\includegraphics[width=\textwidth]{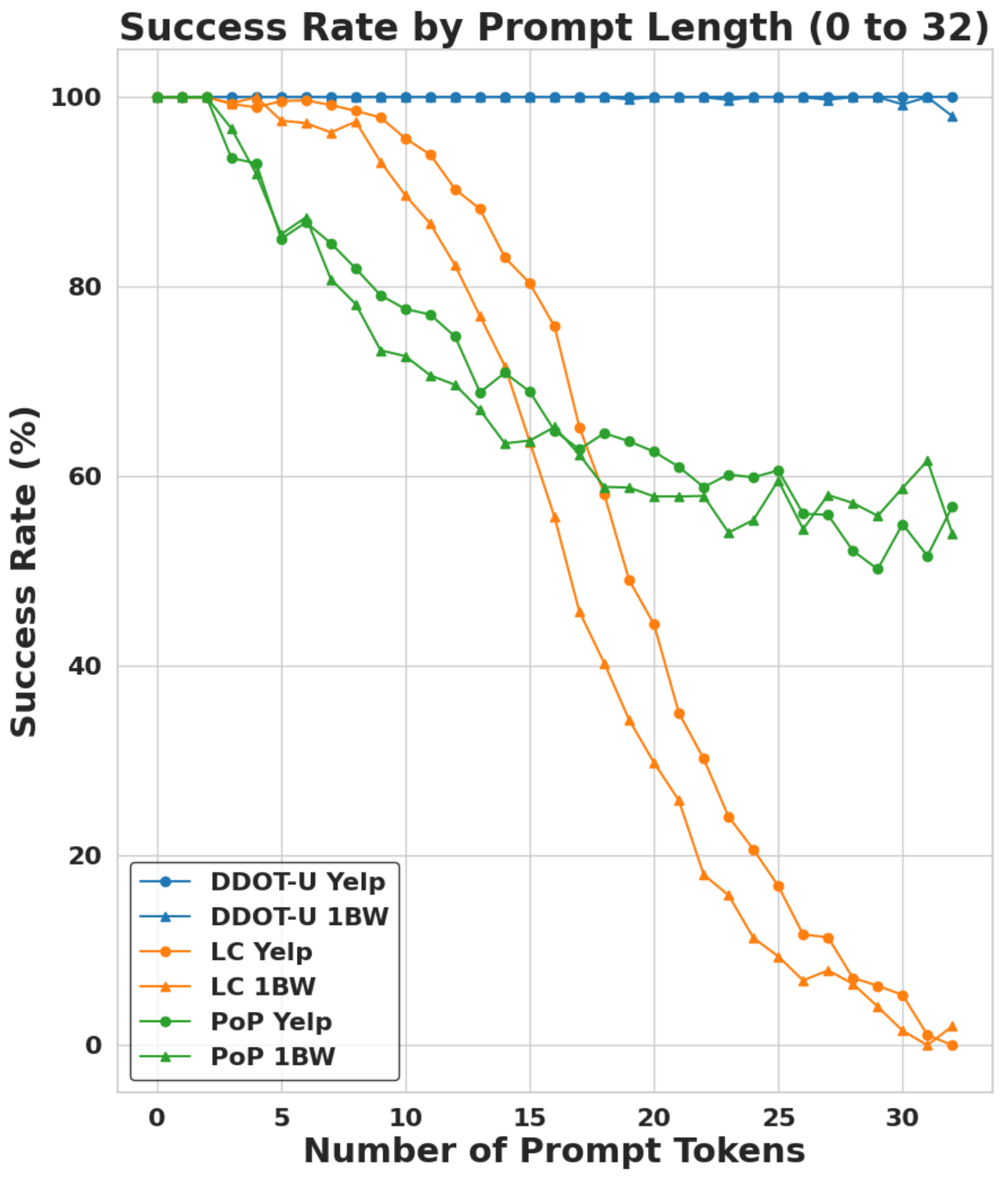}
\vspace{-2.0em}
\captionof{figure}{Success rate on block datasets.  
LC and PoP increasingly generate invalid responses (missing or swapping prompt tokens) as the number of prompt tokens grows.}
\label{fig:num_prompt}
\end{minipage}
\end{table*}

%% file: tables/table3.tex
\begin{table}[!htb]
\centering
\small
\begin{tabular}{l|ccccc}
\hline
\textbf{OT?} & B2 & B4 & N2 & N4 & M \\ \hline
\multicolumn{6}{c}{\textbf{One-Billion-Word}} \\ \hline
No OT & 13.2 & 4.6 & 1.60 & 1.62 & 14.1 \\ 
OT & \textbf{15.7} & \textbf{5.1} & \textbf{3.2} & \textbf{3.2} & \textbf{14.6} \\ \hline
\multicolumn{6}{c}{\textbf{Yelp}} \\ \hline
No OT &  16.39&  5.94&  2.05&  2.10&  16.0\\ 
OT & \textbf{18.6} & \textbf{6.9} & \textbf{3.13} & \textbf{3.21} & \textbf{16.6} \\ \hline
\end{tabular}
\caption{\textbf{Our OT coupling drastically improves preformance across all metrics.} Ablation on OT coupling with small model size.}
\label{tab:OT-Ablation}
\end{table}

%% file: tables/table_new.tex
\begin{table}[ht]
\centering
\resizebox{\columnwidth}{!}{%
\begin{tabular}{lcccccc}
\toprule
\textbf{Model} & \multicolumn{6}{c}{\textbf{Number of Sampling Steps}} \\
\cmidrule(lr){2-7}
 & \textbf{2} & \textbf{4} & \textbf{8} & \textbf{16} & \textbf{32} & \textbf{64} \\
\midrule
\multicolumn{7}{l}{\textit{LC (Left Context)}} \\
Time (s/batch) & 0.71 & 1.43 & 2.86 & 5.72 & 11.5 & 22.9 \\
BLEU-2         & 38.3 & 44.3 & 50.7 & 53.3 & 54.5 & 55.1 \\
BLEU-4         & 27.5 & 33.9 & 41.9 & 45.3 & 47.0 & 47.9 \\
\midrule
\multicolumn{7}{l}{\textit{PoP (Position Prediction)}} \\
Time (s/batch) & 0.53 & 0.90 & 1.64 & 3.13 & 6.09 & 12.0 \\
BLEU-2         & 45.4 & 59.5 & 59.5 & 59.4 & 59.3 & 59.2 \\
BLEU-4         & 40.7 & 50.3 & 50.3 & 50.2 & 50.1 & 50.0 \\
\midrule
\multicolumn{7}{l}{\textit{DDOT-Uniform (Ours)}} \\
Time (s/batch) & \textbf{0.44} & \textbf{0.81} & \textbf{1.54} & \textbf{3.03} & \textbf{5.98} & \textbf{12.0} \\
BLEU-2         & \textbf{59.4} & \textbf{61.2} & \textbf{62.4} & \textbf{63.1} & \textbf{63.5} & \textbf{63.7} \\
BLEU-4         & \textbf{55.5} & \textbf{56.8} & \textbf{57.5} & \textbf{58.0} & \textbf{58.2} & \textbf{58.4} \\
\bottomrule
\end{tabular}%
} 
\caption{\textbf{DDOT achieves superior BLEU scores with faster inference times.} Inference speed (seconds per batch) and BLEU scores on One-Billion-Word for varying numbers of sampling steps. } 
\label{tab:inference_speed} 
\end{table}

%% file: tables/table6.tex
\begin{table*}[ht]
\centering
\begin{tabular}{l|ccccc|ccccc}
\toprule
\textbf{Model} & \multicolumn{5}{c|}{\textbf{One-Billion-Word-Random}} & \multicolumn{5}{c}{\textbf{Yelp-Random}} \\
               & B2   & B4   & N2   & N4   & M    & B2   & B4   & N2   & N4   & M    \\ \midrule
DDOT-R small (Ours)                     & 15.7 & 5.1 & 3.16 & 3.19 & 14.6 & 18.6 & 6.9 & 3.13 & 3.21 & 16.6  \\
DDOT-U small (Ours)                     & 15.5 & 4.9 & 2.94 & 2.97 & \textbf{15.0} & 18.9 & 7.1 & 2.99 & 3.07 & 17.0  \\
DDOT-R medium (Ours)                    & 15.7 & 5.1 & \textbf{3.17}& \textbf{3.21}& 14.6 &  19.7 & 7.0& 3.26 & 3.34 & 17.4  \\
DDOT-U medium (Ours)                     & \textbf{16.3} & \textbf{5.2} & 3.13 & 3.17 & \textbf{15.0} & \textbf{21.2}& \textbf{7.9} & \textbf{3.43}& \textbf{3.52}& \textbf{17.7} \\
\bottomrule
\end{tabular}%
{\caption{Results on One-Billion-Word and Yelp. Metrics are BLEU (B2, B4), NIST (N2, N4), and METEOR (M). Best results are highlighted in \textbf{bold}. \label{tab:results_DDOT-small}}} 
\end{table*}

%% file: tables/table7.tex

\begin{table*}[ht]
\begin{minipage}{\textwidth}
\centering
\small
\begin{tabular}{l|cc|cc}
\toprule
\textbf{Model} & \multicolumn{2}{c|}{\textbf{One-Billion-Word-Random}} & \multicolumn{2}{c}{\textbf{Yelp-Random}} \\
               & D2   & D4   & D2 & D4 \\ \midrule
CBART \cite{CBART}                     & \textbf{49.2} & 82.1 & \textbf{38.1} & \textbf{92.6}  \\
LC   & 37.5 & 89.4 & 21.5 & 69.0 \\
PoP  & 38.85 & \textbf{91.4} & 19.9 & 68.3  \\
DDOT-R (Ours) & 30.7 & 85.3 & 20.2 & 67.7  \\ 
DDOT-U (Ours) & 34.03 & 88.2 & 18.6 & 66.6 \\
\bottomrule
\end{tabular}
\caption{Results on One-Billion-Word and Yelp dataset for generation diversity. Metrics include Diversity (D2, D4). Best results are highlighted in \textbf{bold}.}
\label{tab:results_diversity}
\end{minipage}
\end{table*}

%% file: tables/table8.tex
\begin{table*}[ht]
\centering
\small
\begin{tabular}{l|ccccc}  
\toprule
\textbf{$\lambda$} & B2 & B4 & N2 & N4 & M \\ \midrule
3 & 63.07 & 58.27 &  7.56 & 8.32 & 42.64  \\ 
10 & 63.04 & 58.23 & 7.55 & 8.30 & 42.85 \\
30  & 63.64 & 58.63 & 7.73 & 8.50 & 42.66 \\
100 & 63.05 & 58.18 & 7.56 & 8.32 & 42.6 \\
300 & 63.25 & 58.24 & 7.63 & 8.39 & 42.50 \\
1000 & 63.53 & 58.46 & 7.74 & 8.51 & 42.50 \\
3000 & 62.95 & 57.95 & 7.59 & 8.34 & 42.27 \\
\bottomrule
\end{tabular}
\caption{Impact of different $\lambda$ values on model performance metrics.}  
\label{tab:lambda_values}
\end{table*}

%% file: tables/table9.tex
\begin{table*}[!htbp]
\centering
\small
\begin{tabular}{l|cc} 
\toprule
\textbf{Hyperparameter} & Value \\ \midrule
Batch size & 128 (A30) / 256 (L40) / 512 A(100)\\ 
$\lambda$ & 300 (with scaling) \\
Weight decay & 0.0 \\
Learning rate & $3e-4$ \\
Optimizer & AdamW \\
$\beta_{1}$ & 0.9 \\
$\beta_{2}$ & 0.999 \\
\bottomrule
\end{tabular}
\caption{Training hyperparameter specifications.}  
\label{tab:train_hyperparams}
\end{table*}